\documentclass[10pt,journal,compsoc]{IEEEtran}

\ifCLASSOPTIONcompsoc
\usepackage[nocompress]{cite}
\usepackage[noend]{algorithmic}
\usepackage{caption}
\usepackage{graphicx}
\usepackage{booktabs}
\usepackage{amsmath}
\usepackage{multirow}
\usepackage{amsmath}
\usepackage[ruled,vlined,linesnumbered]{algorithm2e}
\usepackage{ragged2e}
\usepackage{diagbox}
\usepackage{color}
\usepackage{booktabs}
\usepackage{amssymb}
\usepackage{subcaption}

\else
\usepackage{cite}
\fi

\ifCLASSINFOpdf

\else

\fi

\begin{document}

\title{Quantum learning and essential cognition under the traction of meta-characteristics in an open world}

\author{Jin Wang,
        Changlin~Song%
        \IEEEcompsocitemizethanks{\IEEEcompsocthanksitem Jin Wang is with the Department of Computing, Hong Kong Polytechnic University, Hong Kong, China. (sophe\_icon@sina.com)
        \IEEEcompsocthanksitem Changlin Song is with the Department of Computing, Hong Kong Polytechnic University, Hong Kong, China. (charlemagnescl@outlook.com)}%
        \thanks{Manuscript received \today; revised ~~~~~~~~~..}%
}

\markboth{Quantum learning model in an open world}%
{Shell \MakeLowercase{\textit{et al.}}: Bare Demo of IEEEtran.cls for Computer Society Journals}

\IEEEtitleabstractindextext{%
\justifying\let\raggedright\justifying
\begin{abstract}
Artificial intelligence has made significant progress in the Close World problem, being able to accurately recognize old knowledge through training and classification. 
However, AI faces significant challenges in the Open World problem, as it involves a new and unknown exploration journey. 
AI is not inherently proactive in exploration, and its challenge lies in not knowing how to approach and adapt to the unknown world. 
How do humans acquire knowledge of the unknown world? Humans identify new knowledge through intrinsic cognition. 
In the process of recognizing new colors, the cognitive cues are different from known color features and involve hue, saturation, brightness, and other characteristics.
When AI encounters objects with different features in the new world, it faces another challenge: where are the distinguishing features between influential features of new and old objects? 
AI often mistakes a new world's brown bear for a known dog because it has not learned the differences in feature distributions between knowledge systems. 
This is because things in the new and old worlds have different units and dimensions for their features.
This paper proposes an open-world model and elemental feature system that focuses on fundamentally recognizing the distribution differences in objective features between the new and old worlds.
The quantum tunneling effect of learning ability in the new and old worlds is realized through the tractive force of meta-characteristic.
The outstanding performance of the model system in learning new knowledge (using pedestrian re-identification datasets as an example) demonstrates that AI has acquired the ability to recognize the new world with an accuracy of $96.71\%$ at most and has gained the capability to explore new knowledge, similar to humans.
\end{abstract}

}

\maketitle

\IEEEdisplaynontitleabstractindextext

\IEEEpeerreviewmaketitle

\IEEEraisesectionheading{\section{Introduction}\label{sec:introduction}}
\IEEEPARstart{I}{n} the world we live in, the unknown always surpasses the known, and open-world scenarios are prevalent. 
In the development of artificial intelligence, significant progress has been made in learning about the known world, particularly in cognition and classification tasks of known categories, assuming a closed-world setting. 
However, exploring and dealing with the unknown world is the direction for further advancement in artificial intelligence.
In this situation, the traditional closed-world assumption is no longer applicable, and the concept of the open world is introduced to address this challenge.

The gap between the known world and the unknown world creates a barrier in knowledge and data distribution, hindering the progress of AI exploration.
AI's active learning in the unknown world is often based on predefined exploration strategies and is constrained by the limited knowledge and unknown data distribution in that world. 
Although domain adaptation techniques suggest that AI can explore the world based on knowledge transfer, it still fundamentally relies on the assumption of using the known to infer the unknown.

Therefore, our research focuses on measuring the gap between the known and the unknown, determining the unknown, and labeling it. 
The first aspect to study is how to evaluate and measure the unknown. 
This evaluation standard goes beyond shallow approaches that calculate distances (such as Euclidean or cosine) between the unknown and the known based on known features.
The deeper differences need to be linked to the definition of the scale, specifically identifying which common characteristics between the unknown and the known are worth measuring to ultimately distinguish the unknown. 
We are attempting to establish a magnetic channel between the known and unknown, capable of generating quantum entanglement to tightly connect the characteristic states of the two worlds and unleash the potential of superlearning. Enabling the learning ability to exhibit quantum tunneling effects through the channel of essence of laws and cognition.

Based on this motivation, when designing our models, we focus on enabling them to learn the common features between the unknown and the known. 
We aim to facilitate the exploration of measurement standards and adaptation to the unknown world from the beginning, promoting the improvement of the model's capabilities through the interaction between the known and the unknown.
Furthermore, the way a model approaches the unknown world also determines its ability to assimilate knowledge and incorporate new knowledge gained from exploring the unknown. 
Depending on how a model handles unknown categories, there are two main problems: the Open Set problem and the Open World problem, which were introduced by Scheirer et al.\cite{f44} and Bendale et al.\cite{f45}, respectively. 
The former requires the model to be able to recognize and reject samples from unknown categories, while the latter demands that the model is capable of classifying or labeling samples from unknown categories.
In theory, Open World models have lower accuracy in recognizing and classifying compared to Closed World models\cite{f45}.
Also the Open World problem imposes certain requirements on the sampling of data samples.
However, the Open World approach requires the model to possess the ability to enhance itself through exploration and to engage in continuous learning.

Compared to well-established IDE networks, the significance of Open World models lies in exploring new domains. 
No matter how fine-grained or accurate the classification of classifiers is, it is still a limitation imposed by the definition of knowledge at its source. 
In the vast world, our way of cognition is to start from the known world while experiencing the unknown world. 
Through the quantum tunneling effect between the known and unknown, we can identify and learn more, transferring knowledge and learning abilities.

As the foundation of the quantum learning, the main model needs to support simultaneous learning and execution, acquiring new knowledge through comparison and feature unification, annotating and continuously iterating on the new knowledge.
We introduce an online learning method, which involves continuously receiving new data in the new world and conducting incremental learning to update the model based on its pre-trained state.
To identify the common features and measure the differences between known and unknown aspects, we introduce a model based on autoencoder to extract shared features. 
We propose a knowledge evaluation method to measure the gap between the known and unknown aspects based on these shared features.
To ensure that the model explores new knowledge while not forgetting the previously acquired knowledge, we adopt a semi-supervised learning approach with new and old knowledge learning in one sharing model. 
This allows the model to establish a special interdependence between the quantum subsystems of the new and old worlds, enabling the transmission of information and the formation of quantum channels and entanglement.

In the context of designing a meta-characteristic-guided system that adapts to these fundamental features, the concept of "quantum tunneling" is used as a metaphor. 
Quantum tunneling refers to the phenomenon in quantum mechanics where particles can pass through energy barriers that would be impossible in classical scenarios. 
In the context of the learning systems of the classical and quantum worlds, this means the ability to traverse or explore different states or feature spaces in a non-deterministic manner and overcome knowledge transfer barriers between the old and new systems through the exchange of meta-features in the system.

The challenges faced by the open world are as follows:
(1) In the new and old worlds, things have different units and dimensions for their features, and there is no universal standard to measure the general features of things for cognition.
There is no universally recognized standard for measuring the general characteristics of things in order to identify them.
(2) There are significant differences in data distribution and feature representation between different domains. 
An open-world model trained on one domain cannot be directly applied to another domain, as its generalization ability is poor.
(3) When the model learns new knowledge, comparing and integrating new and old knowledge face the problem of insufficient confidence in the new knowledge. 
Insufficient feedback on new knowledge leads to imbalanced data samples.
(4) When using an open world model for incremental recognition, there is a challenge of an infinite label space, which brings challenges such as imbalanced old and new training samples, computational complexity, and the risk of forgetting knowledge.
\begin{figure}
\centering
\includegraphics[scale=0.261]{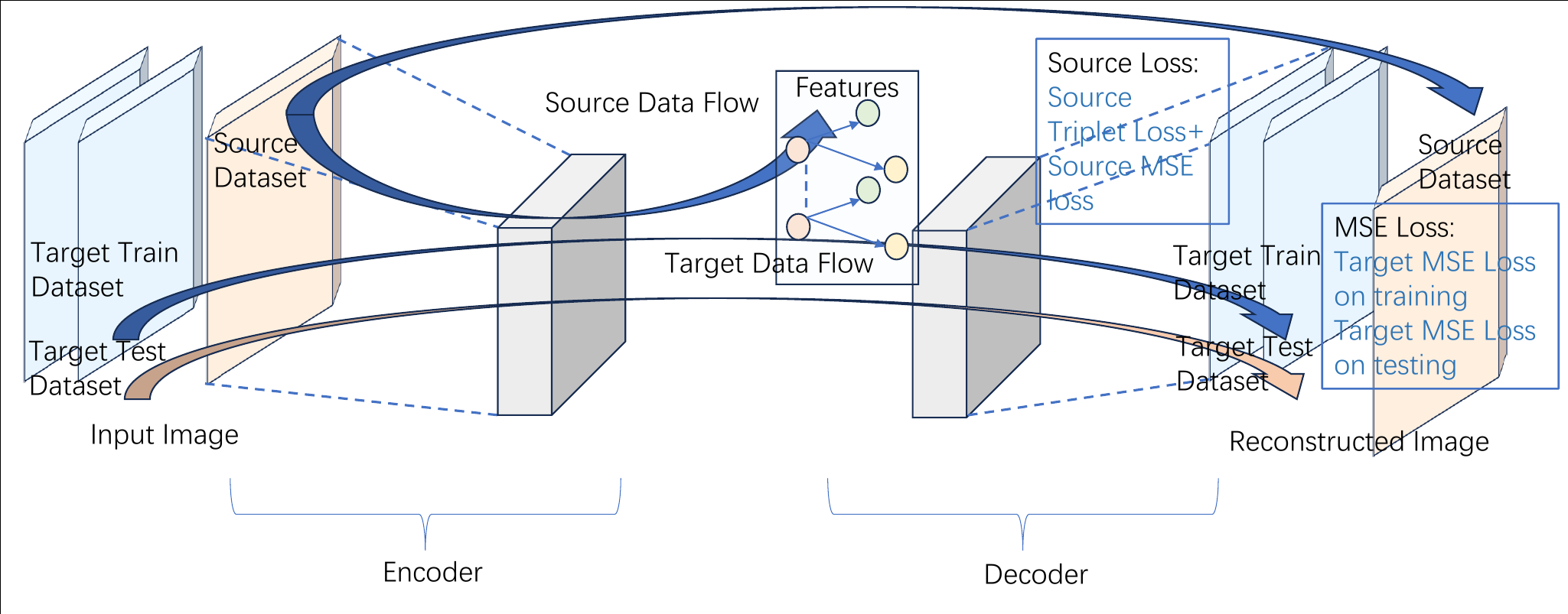}
\caption{Semi-supervised learning model}
\label{figure01}
\end{figure}
In summary, the main contributions of this paper are as follows:
\begin{itemize}
  \item[\textbullet]
  We have designed a learning system using quantum concepts that enables the learning capability to overcome knowledge transfer barriers between the old and new systems in the classical and quantum worlds. 
  This system achieves the phenomenon of quantum tunneling.
  \item[\textbullet]
  We propose Meta-features as the representations of essential cognition, which are driving advancements in the field of New World Learning Systems as the learning traction.
  \item[\textbullet]Based on the semi-supervised learning autoencoder, the open world model as illurstrated in Fig.\ref{figure01} is introduced, incorporating online learning methods, where the unidentified target domain data is continuously involved.
This allows the model to continuously adapt to the learning rules of the unknown world.
  \item[\textbullet]Semi-supervised learning is introduced to enable training on a single model, providing the possibility to compare and fuse new and old knowledge.
  \item[\textbullet]Taking into account the iterative collision of new and old knowledge in the evaluation, we propose knowledge evaluators from the source and target domains that engage in synchronous optimization and communication, constituting information exchange within the quantum channel.
  \item[\textbullet]We propose an open-world learning framework consisting of the main model and two knowledge evaluators. 
  In the model, through the comparison, competition, supplementation, and unification of the known and unknown aspects, a sufficient barrier and energy have been formed. 
  Under the guidance of meta-features, we have triggered the quantum tunneling effect.
\end{itemize}

\section{Related Works}\
In the traditional closed-world classification assumption, classification models only consider known classes and assume that test samples belong to one of these known classes. 

Many studies have introduced low-confidence pseudo-labels for unsupervised learning to incorporate unlabeled target domain data into the training process, such as in the cited references \cite{f12, f4, f14}. 
These methods utilize unlabeled data for training and attempt to classify unknown classes.
Some methods utilize unsupervised learning to generate robust person images, overcoming the label scarcity problem in classification tasks, as mentioned in the cited reference \cite{f30}.
Given the challenges of different data distributions, insufficient training data in the target domain, and differences in feature representation in re-identification tasks, transfer learning \cite{f7} and domain adaptation methods \cite{f11} have been introduced to adjust models for better classification in open-world scenarios.

The introduction of the meta-learning problem for quantum computers aims to explore how to enhance the efficiency and performance of the learning process in a quantum computing environment. 
It investigates the application of meta-learning principles in quantum computing to improve the efficiency and performance of learning processes on quantum devices\cite{f48}. 
The referenced literature\cite{f49} provides new insights and methods for addressing complex quantum learning tasks\cite{f49}.
However, to date, no work has enabled AI models in an open-world environment to achieve unparalleled learning capabilities through quantum properties.

The Open World Model is a learning approach designed to handle open-world classification problems. 
OpenMax is a method specifically developed for open-world classification, improving classification models by modeling unknown classes in the open world \cite{f46}. 
A large-margin-based approach for open-world classification is proposed in the cited reference \cite{f48}. 
G-OpenMax is another open-world classification method based on Generative Adversarial Networks (GANs) \cite{f49}.
These methods aim to improve the ability of classification models to handle unknown classes, adapting to the challenges of open-world classification tasks. 
They leverage unsupervised learning, transfer learning, and GANs techniques to enhance the performance of classification models.

In the field of person re-identification, domain adaptation techniques\cite{f11} have an advantage over transfer learning\cite{f7}, \cite{f10} due to significant inter-domain distribution differences. 
Transfer learning involves pre-training on a source dataset and fine-tuning on a target dataset, enabling the model trained on the source dataset to adapt to the target scenario\cite{f23},\cite{f5}. 
However, in an open-world setting, due to variations in environmental conditions and camera setups, a model trained on one dataset may perform poorly on another dataset. 
Domain adaptation techniques aim to bridge this gap and improve the model's generalization capability\cite{f24} but it addresses the smaple lacking problem not a openworld problem.

To incorporate unlabeled target domain data into the training process, most subsequent research introduced pseudo-labeling for achieving unsupervised learning, such as \cite{f12}, \cite{f4}, \cite{f14}. 
The core idea of \cite{f12} is to establish a universal feature representation by jointly training data from multiple domains to achieve cross-domain person re-identification. 
\cite{f14} proposed a method of soft multi-label learning, which assigns multiple labels to target domain samples instead of traditional single labels, capturing the diversity and uncertainty of the samples.
However, pseudo-labeling often introduces errors.

When evaluating new and old knowledge, we often employ distance measurement methods.
Because this method provides a general assessment, combining more context and features, we introduce a knowledge evaluation model for a more reliable comprehensive judgment.
We introduce a triplet to evaluate the association between new and old knowledge from the perspective of inter-class distance.
Triplet Loss is based on a simple and direct idea: minimizing the intra-class distance while maximizing the inter-class distance, playing an increasingly important role in metric learning\cite{f8}. 
Batch images are used to construct many triplets, and the triplet loss effectively drives the model to learn better feature representations, making the anchor point more similar to the positive point and less similar to the negative point.
In summary, Triplet Loss can optimize the similarity between the anchor point and positive sample to be higher than the similarity between the anchor point and negative sample\cite{f31}. 
Some research\cite{f32} proposes selecting the most easily misclassified samples from the negative samples to improve the discriminative ability of the model. 
The paper\cite{f9} suggests mining challenging triplet samples from the training data, which also means using multiple candidate classes to improve the robustness of the model.
The comparison between the similarity distance difference between the anchor point and the nearest negative sample and the farthest positive sample and the margin is not sufficient to determine whether the anchor point belongs to a candidate class.

Quantum learning is an emerging subfield within the field of quantum information that combines the speed of quantum computing with the learning and adaptation capabilities offered by machine learning,
Quantum learning can be divided into two parts:
1. Machine learning algorithms that use data obtained from quantum computers and traditional computers as training data to train models\cite{f50}.
2. Improving machine learning algorithms through ideas and models of quantum mechanics is also part of quantum machine learning.
In the quantum learning system of the old and new world, the fundamental features are characterized by uncertainty and probability distributions. 
Similar to the nature of quantum phenomena, we cannot know the precise values of these features with certainty, but we can only obtain their probability distributions through measurements or observations. 
This is the basic feature of quantum learning.
The relationships between these features may exhibit the characteristics of quantum entanglement, where the state of one feature can instantaneously affect the state of another, regardless of their spatial separation.
In the design of our meta characteristic traction system, we have incorporated this fundamental feature and achieved the quantum tunneling effect through the traction formed by the comparison, aggregation, and collision of meta characteristics.
\section{Methods}\
Assuming the data and labels on the source dataset $\mathcal{X_s}$ are represented as data $x_i$ and label $z_i$, and the target dataset $\mathcal{D}$ as $D={d_i}$, where $i$ is the number of samples.
Suppose we have an autoencoder with input data $x$, encoder output $y$, and decoder output $x'$. 	
We can represent the encoder as a function $g(x)$, and the latent feature of the data is denoted as $y_i = g(x_i) \in \mathbb{R^m}$. 
where $\mathbb{R^m}$ represents the feature spaces to which the input data $x_i$ is mapped.
The metacharacteristic is the extraction of metric-invariant feature B from feature A, where feature A can be measured using feature B. 
It captures certain important attributes or characteristics of feature A and enables evaluation through comparison in both new and old contexts. 
The goal is to learn a function $f$ that extracts features $B(i)$ with metric invariance from the features $A_i$ of the sample $x_i$, with $E(i) = f(A_i) \in \mathbb{R^n}$, where $\mathbb{R^n}$ represents the $n$-dimensional meta characteristic feature space to which the feature $A_i$ is mapped.
This metric invariance holds across different feature sets.
\begin{figure*}[htp]
\centering
\includegraphics[scale=0.51]{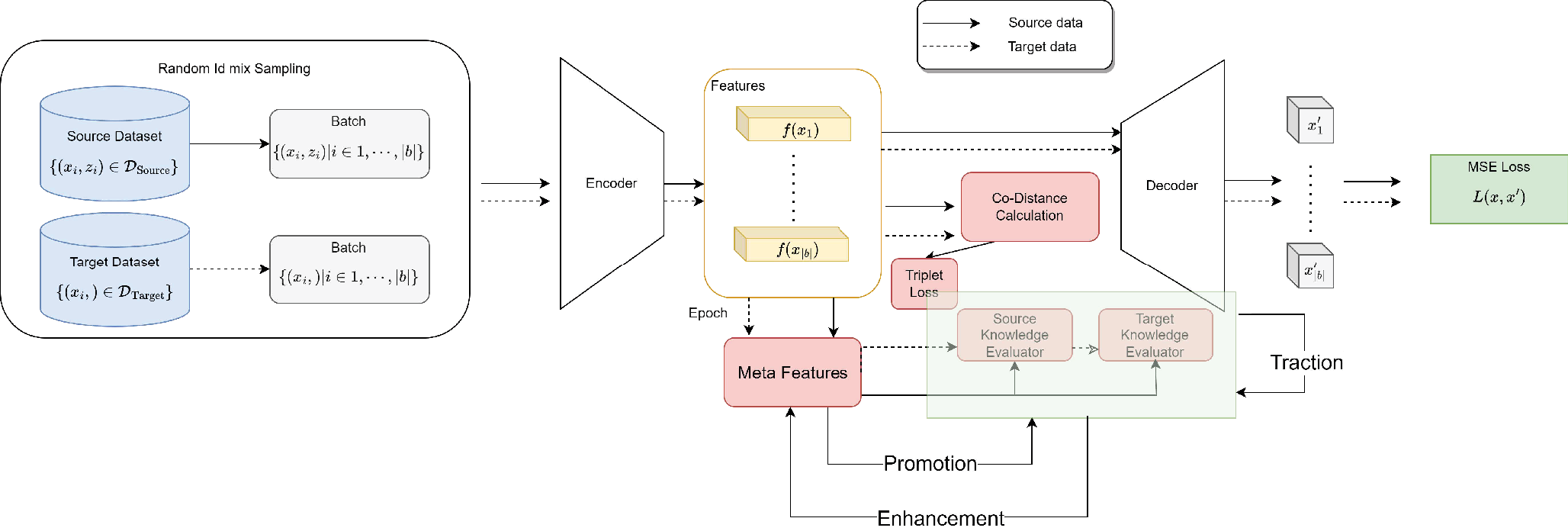}
\caption{Training stage}
\label{fig:system}
\end{figure*}
The reason we adopt a semi-supervised learning model as shown in \ref{fig:system} is that in the new knowledge system, there is a scarcity of data and low cognitive confidence. 
To maximize the utilization of data from the old knowledge system, we leverage the information from both labeled data in the old knowledge and unlabeled data in the new dataset through a comparative and symmetrical approach. 
This architecture provides the possibility of establishing quantum channels within different quantum subsystems. 
\begin{itemize}
  \item[\textbullet] Main Model Training Phase: The update process of the source domain loss in the main model consists of two components: the Triplet Loss calculated on the source domain data through the encoder and the MSE Loss calculated on the source domain data through the autoencoder.
Our model supports online learning, where the MSE Loss is updated when testing the target domain data on the autoencoder.
On one hand, this approach helps alleviate the problem of insufficient target domain samples.
On the other hand, the intersection and collision of new and old knowledge enhance the model's generalization ability and promote its evolution.
  \item[\textbullet] Main Model Testing Phase: Our model can continuously update the target domain loss and learn during the testing phase as well. 
  While being re-identified, it calculates the target domain MSE loss and updates accordingly, maintaining the ability for online learning.
  \item[\textbullet] Knowledge Evaluator Training Phase: The initial learning samples for the knowledge evaluator are the characteristics rlated to feature distance of the source domain. 
  As the testing in the new knowledge domain progresses, when encountering confident target domain characteristic parameters related to feature distance and discriminatory results, they are provided for training and learning by the knowledge evaluator. 
  This enables the knowledge evaluator to make better judgment decisions by incorporating more target domain information in conjunction with the knowledge from the source domain.
  \item[\textbullet] Knowledge Evaluator Testing Phase: The logistic regression predicts the label based on the characteristic related to feature distance of the tested image and the nearest distance to a candidate class.
  This testing is only applicable to the target domain.
\end{itemize}
\subsection{Knowledge evaluator}\
Humans learn new knowledge by starting from known categories of knowledge, understanding the essential characteristics of that category, and then analyzing the essential features of new knowledge from the same perspective. 
By measuring the differences in features between new and old things and referring to the most similar known objects, they determine whether the new thing belongs to the same category as the old knowledge. This is a universal law of cognitive exploration.
In this process, the reference of the knowledge evaluation system should be as close to the essence as possible, and the knowledge evaluation system itself should preferably be evolutionary. 
On the other hand, discrimination conclusion should be as simple as possible, only requiring a decision of yes or no.

Existing open-world models such as NCM and NNO\cite{f45} use fixed parameters that have been trained for category measurement, and the category measurement is not optimized in the new world. 
The measured feature parameters exist in two different dimensional spaces, one for the unknown world and the other for the known world, leading to imbalanced feature distributions. 
Previous work using a single metric suffers from performance degradation due to scale differences.
Our model sets up two knowledge evaluators for new and old knowledge respectively, breaking free from the constraints of the old knowledge space in the exploration of the new world.
The metric learning in NCM and NNO is limited to the close-set of known object categories and falls short in exploring the subsequent unknown world.
Our knowledge evaluators can continuously optimize in the constantly acquired new knowledge.
The two discriminators in the feature space of the source-target domain of our model use the same meta-features to evaluate knowledge, maintaining synchronized communication for the essence understanding of things.
The Knowledge Evaluator as shown in Figure \ref{fig:system} is a system for knowledge evaluation and discrimination.
\subsection{Source and Target evaluations}\
We adopt a method of training the feature evaluator of the new and old knowledge systems through the exchange of meta-characteristic data from the source domain training set and the target domain validation set, in order to facilitate the continuous optimization of the feature evaluator.
As shown in Fig. \ref{fig:system}, The use of different evaluators ensures the optimization of distinguishing essential features between the old and new knowledge domains. 
The data crossover facilitates knowledge exchange between the two environments, promoting mutual optimization.
\begin{figure*}[htp]
\centering
\includegraphics[scale=0.50]{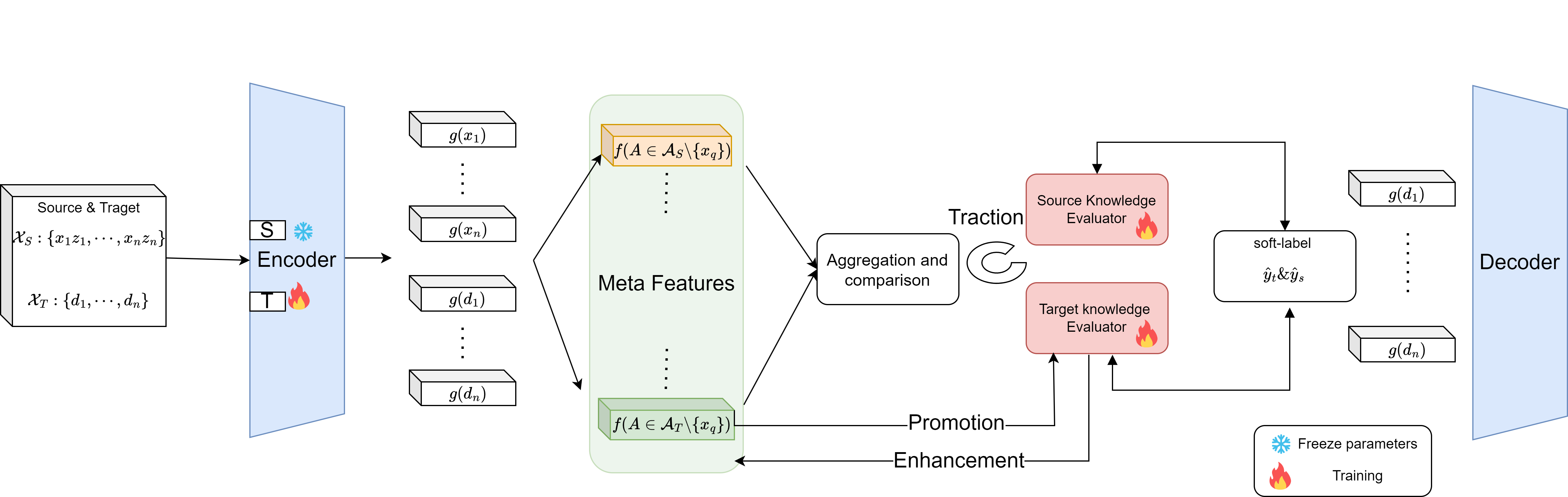}
\caption{Knowledge evaluator: Querying Stage}
\label{figure03}
\end{figure*}
\begin{itemize}
  \item[\textbullet]
  Traction: The traction of the knowledge evaluator towards the cognitive understanding of the new world comes from the communication between the metric systems $R_m$ and $R'_m$ in the metacharacteristic spaces of the old and new worlds, regarding the soft labels (identification) $\hat{y}_t$ and the true labels $\hat{y}_s$. 
  The coefficient of variation (CV) eliminates the uneven distribution between the feature spaces $R_E$ and $R'_E$. 
  Through the aforementioned communication, the old knowledge system provides experiential reference to the cognitive understanding of the new knowledge system, thereby guiding and accelerating the learning of the new knowledge system.
  \item[\textbullet]
Enhancement: The metric systems $R_m$ and $R'_m$ of the old and new worlds receive feedback from the soft labels $\hat{y}^t$ and true labels $\hat{y}^s$ of the training and testing sets. 
The metacharacteristics $R_m$ and $R'_m$ measure the feature spaces $R_E$ and $R'_E$, respectively. They are used by the evaluators to assess the changes in the feature spaces $R_E$ and $R'_E$.
 Statistical indicators such as mean, variance, etc., extracted from the metacharacteristics, enhance the evaluation criteria of the knowledge evaluator in the new world from multiple perspectives. 
 They also strengthen the measurement capabilities of the metacharacteristics in the old world after the evaluator's communication with the new world.
  \item[\textbullet]
Promotion: Since the deformation measurement parameter $mycv$ can measure the symmetric features in the old and new knowledge systems, assuming that the metric systems $R_m$ and $R'_m$ of the old and new worlds obtain the asymmetric difference $\delta R_n$ by comparing the measured feature spaces $R_E$ and $R'_E$, it can be mapped to the metacharacteristic space $\delta R_m$ through a kernel function. 
After sufficient communication between the knowledge evaluator and the evaluation results of the old and new world knowledge, combined with their respective state transitions, the evaluation criteria are improved, promoting the transfer of experience between the old and new worlds.
\end{itemize}
This is an online learning system because during the testing process, the MSE loss of the test set is continuously used to update the model, and the soft labels also facilitate the evolution of the knowledge evaluator.
The knowledge evaluator's understanding of the representation significance of the constantly changing features and metacharacteristics, as well as the mapping between the feedback soft labels, enables the evolution of decision-making and provides more room for parameter selection in the system.
\subsection{Meta characteristic}\
Meta characteristic data refers to features that describe other features. 
It can include statistical indicators in the feature space, correlations between attributes, and the distribution of data.
We define statistical indicators such as the mean and variance between feature values, and measure the correlation coefficients between features, such as variance.
Specifically, we introduce a normalized soft indicator called the coefficient of variation.
1.Coefficient of Variation of AP (distance of anchor-positive)
\begin{equation}
CV_{AP} = \frac{{\sigma_{AP}}}{{\mu_{AP}}} \times 100
\end{equation}
where standard deviation is denoted by $\sigma_{AP}$, $\mu$ is the mean value, and $AP$ is denoted as the feature distance between the input image to be recognized and the positive samples of the candidate class.
2.Coefficient of Variation of AN (distance of anchor-negative)
\begin{equation}
CV_{AN} = \frac{{\sigma_{AN}}}{{\mu_{AN}}} \times 100
\end{equation}
The new knowledge system and the old knowledge system have different units or dimensions. 
The coefficient of variation, which standardizes the mean, is a method to eliminate the uneven distribution of feature sizes. 
It can adapt to the characteristics of different class distances in the new and old worlds.
where standard deviation is denoted by $\sigma_{AN}$, and AN is denoted as the feature distance between the input image to be recognized and the negative samples of the candidate class.
We also define a deformation based on the coefficient of variation:
\begin{equation}
mycv = (CV_{AP} - CV_{AN}) / (CV_{AP} + CV_{AN})
\end{equation}
This metric method normalizes the difference values within the range of -1 to 1, regardless of the absolute numerical values. 
This makes the comparison of differences between different feature data fairer and more reliable.
The formula for the coefficient of variation in the new and old knowledge systems, $(a - b) / (a + b)$, is a normalization of the coefficient of variation between two domains. 
It calculates the relative difference between the coefficient of variation. 
Moreover, because this formula exhibits symmetry, it is well-suited for capturing symmetric features present in the new and old knowledge systems, effectively transferring the learning experience from the old knowledge system to the new one.
Due to its high robustness, this flexible parameter can adapt to the measurement requirements of different fields. 
It allows for a flexible comparison and enhancement of similarities and differences in the new and old worlds.
\subsection{Quantum leanrning}\
\begin{figure}
\centering
\includegraphics[scale=0.319]{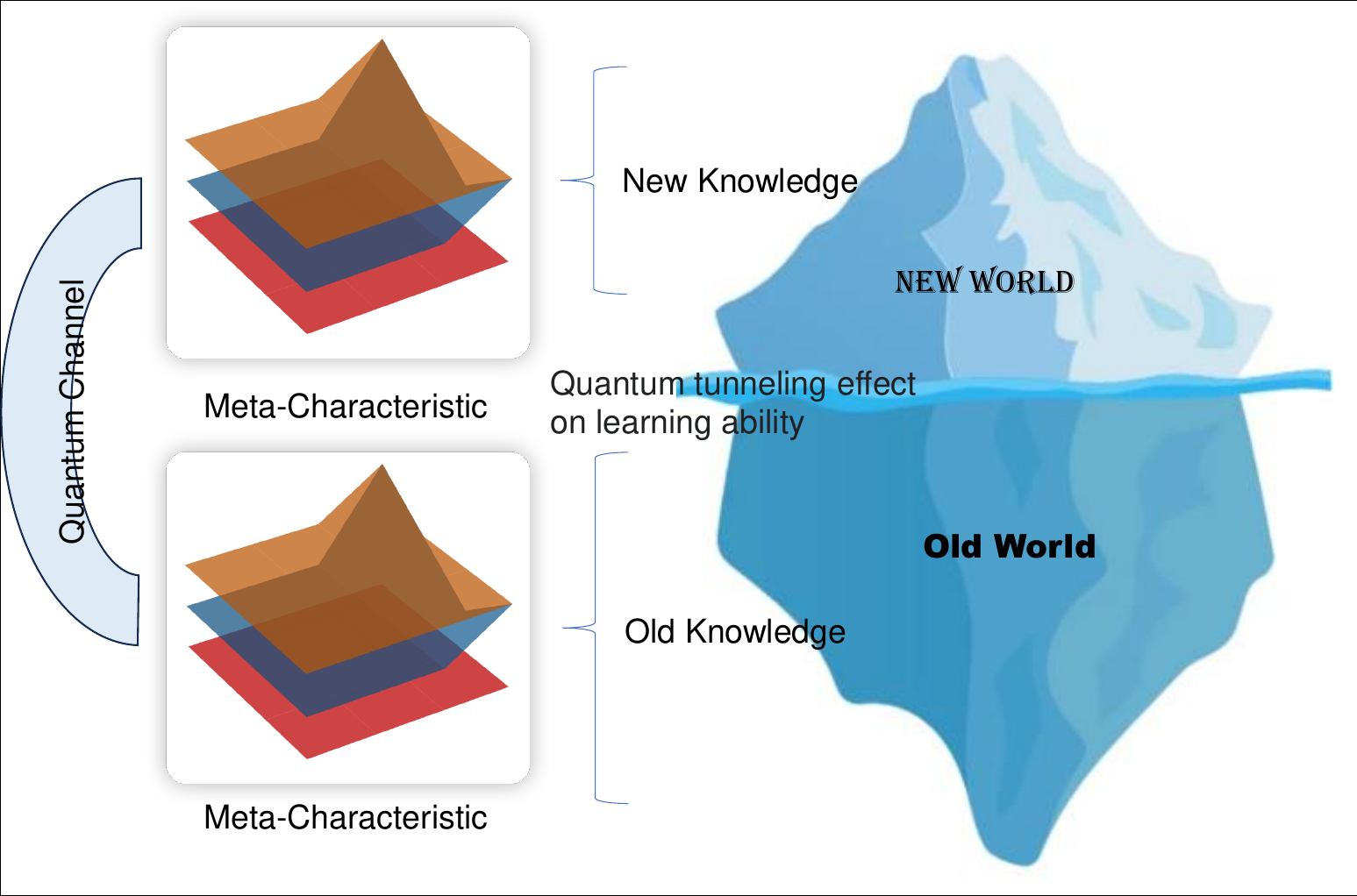}
\caption{Quantum Learning and Quantum Tunneling Effect of Learning ability}
\label{fig:quantum}
\end{figure}
As shown in Figure \ref{fig:quantum}, different classifiers ensure the differentiation and optimization of the essential features of the old and new worlds, while data intersections ensure the exchange of knowledge between the two environments to promote co-optimization.
High-dimensional elemental features serve as the outermost encrypted channel of the quantum channel, simultaneously shielding the escape of inner-layer data.
Meta-characteristic is the driving force behind the quantum channel covered by elemental features.
The traction of the meta-characteristic drives the continuous transfer of learning ability through the channel between the new and old worlds, witnessing the quantum tunneling effect of learning ability in the new and old worlds.
The occurrence of quantum tunneling effect is accompanied by the measurement of quantum entangled states, and this effect will be demonstrated through experiments.

\section{Performance Evaluation}\
We evaluate the proposed model on two popular ReID(Re-identification) datasets in computer vision field: Market-1501 and DukeMTMC-Re-ID.
A mixed sampling strategy was employed, including random sampling from the same ID and random sampling from different IDs. 
This sampling method is much closer to the real distribution of a completely unfamiliar new world compared to the conventional approach of sequentially sampling from a gallery set, which is defined as openworld dataset.
Furthermore, our model completely disregards the classification label from the old world, meaning that we do not utilize the classification label obtained from training on the old world.
Our proposed model focuses on determining whether an image belongs to a candidate person ID class. 
In this case, there is no need to evaluate the performance of the person re-identification network using the mAP (mean Average Precision) metric at different thresholds when the Rank-1 accuracy is very high.	
ReID-Rank1 Accuracy of the same class: statistical ratio in one epoch when the input image to be re-identified belongs to the candidate class (the nearest class in feature distance) with the unit: $\%$.
ReID-Rank1 Accuracy of the different class: statistical ratio when the input image to be re-identified does not belong to the candidate class.

\subsection{Experiment evaluation}\
\begin{figure*}[htb]
  \centering
  \begin{subfigure}[b]{0.23\linewidth}
     \begin{center} 
     \includegraphics[scale=0.32]{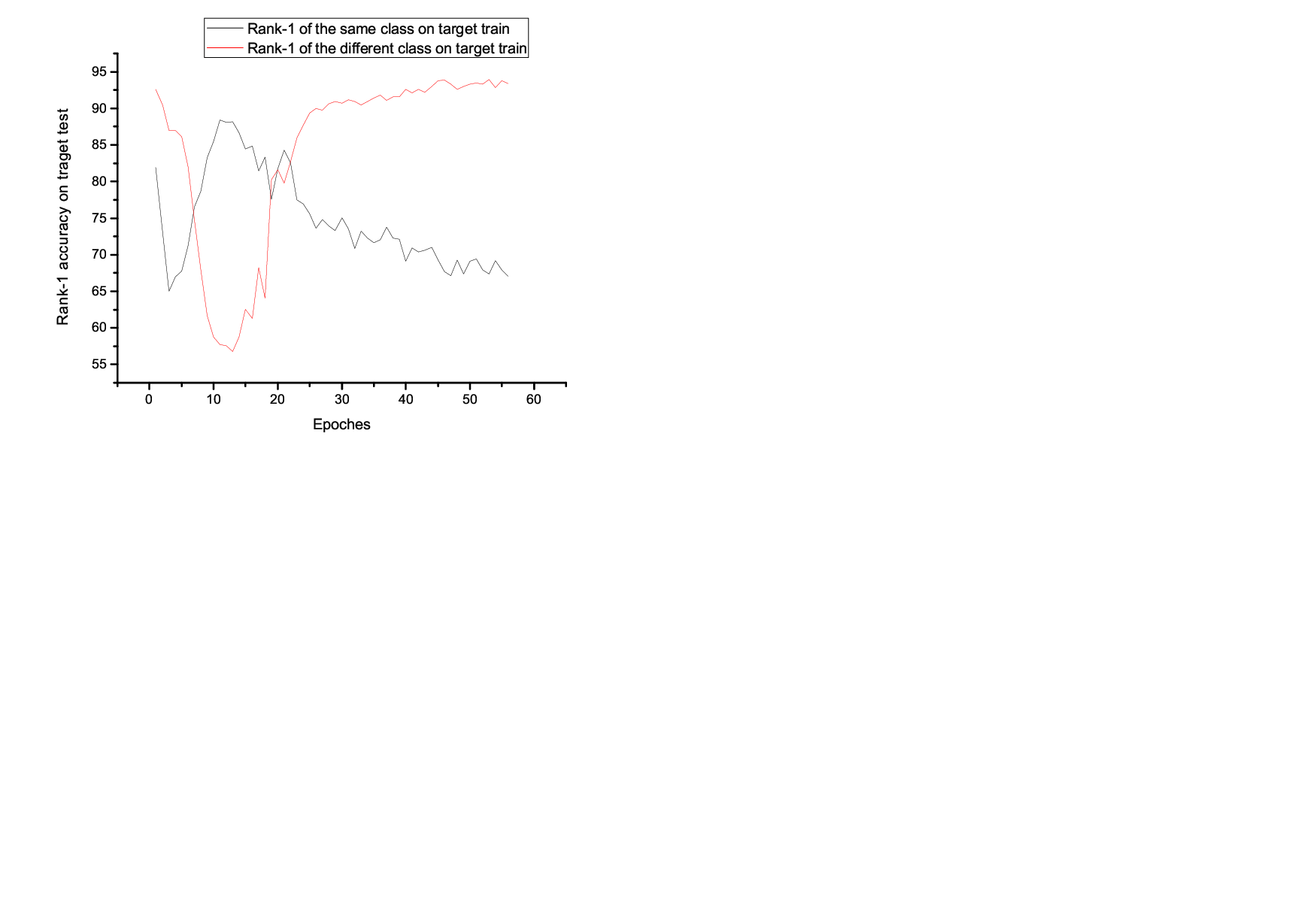}
     \vspace{-4cm}
    \caption{Target train}
    \label{fig:a}
    \end{center} 
  \end{subfigure}
  \hfill
  \begin{subfigure}[b]{0.23\linewidth}
     \begin{center} 
     \includegraphics[scale=0.32]{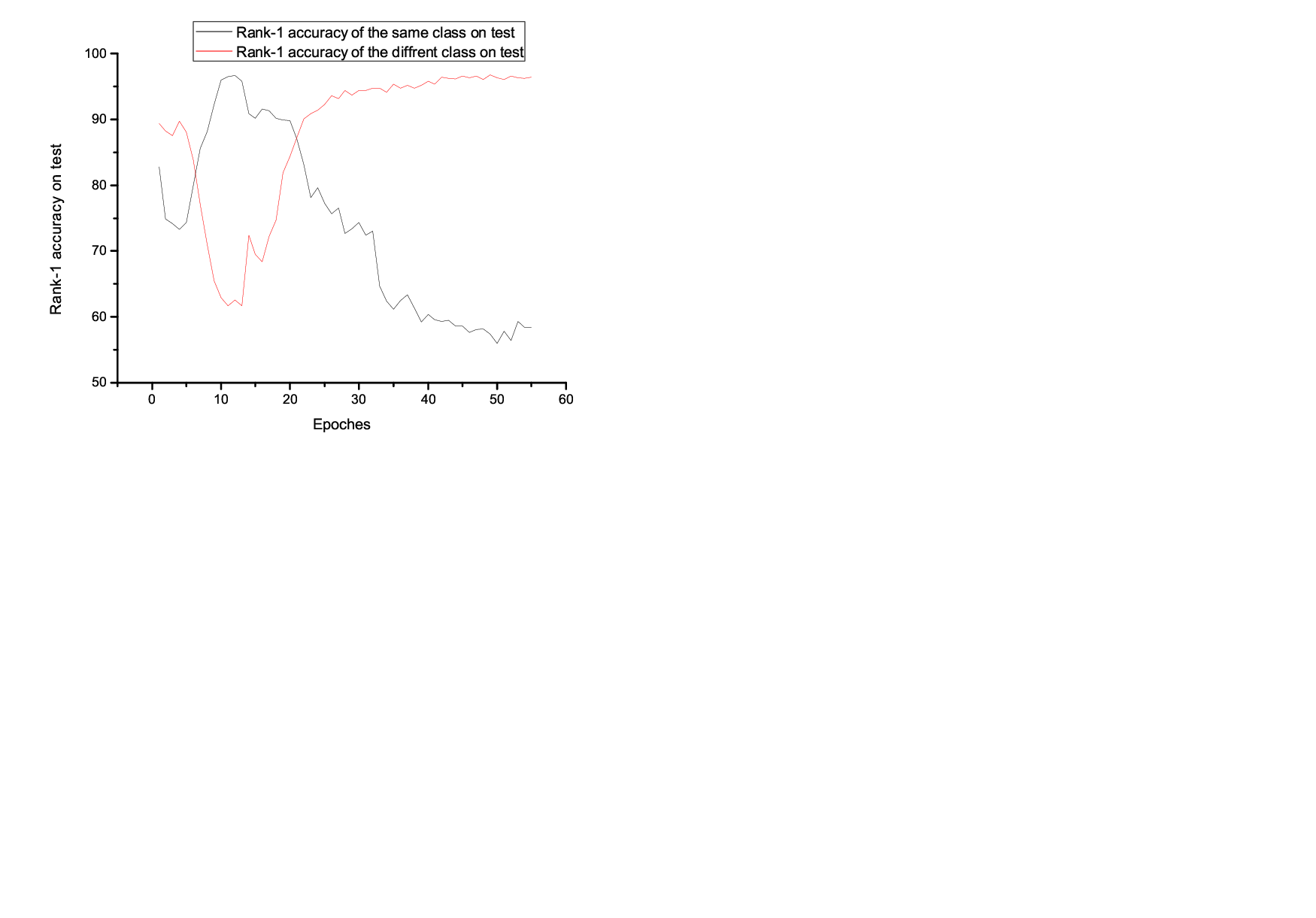}
     \vspace{-4cm}
    \caption{Target test}
    \label{fig:b}
    \end{center} 
  \end{subfigure}	
    \hfill
  \begin{subfigure}[b]{0.23\linewidth}
     \begin{center} 
     \includegraphics[scale=0.32]{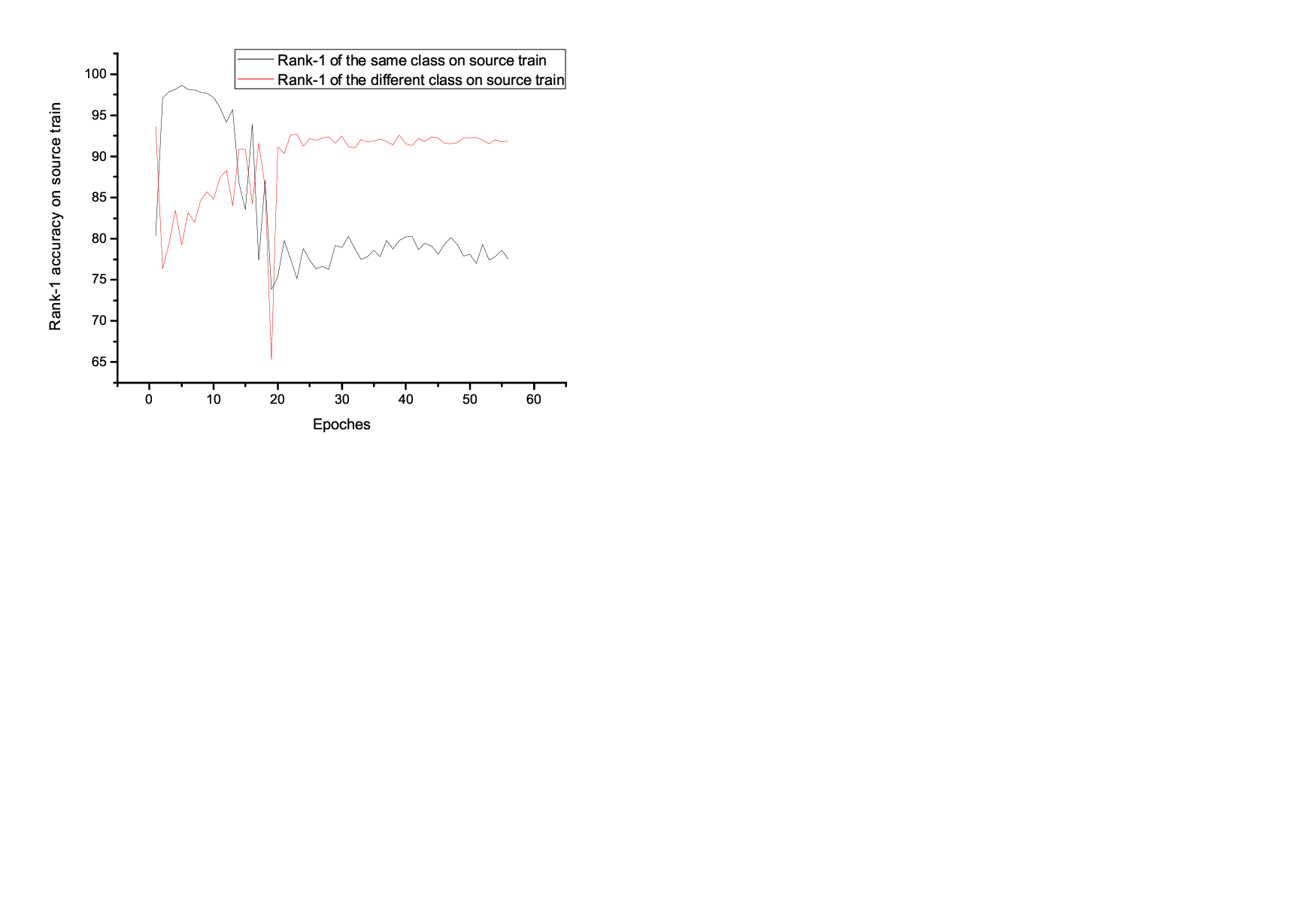}
     \vspace{-4cm}
    \caption{Source train}
    \label{fig:c}
    \end{center} 
  \end{subfigure}
  \hfill
  \begin{subfigure}[b]{0.23\linewidth}
     \begin{center} 
     \includegraphics[scale=0.32]{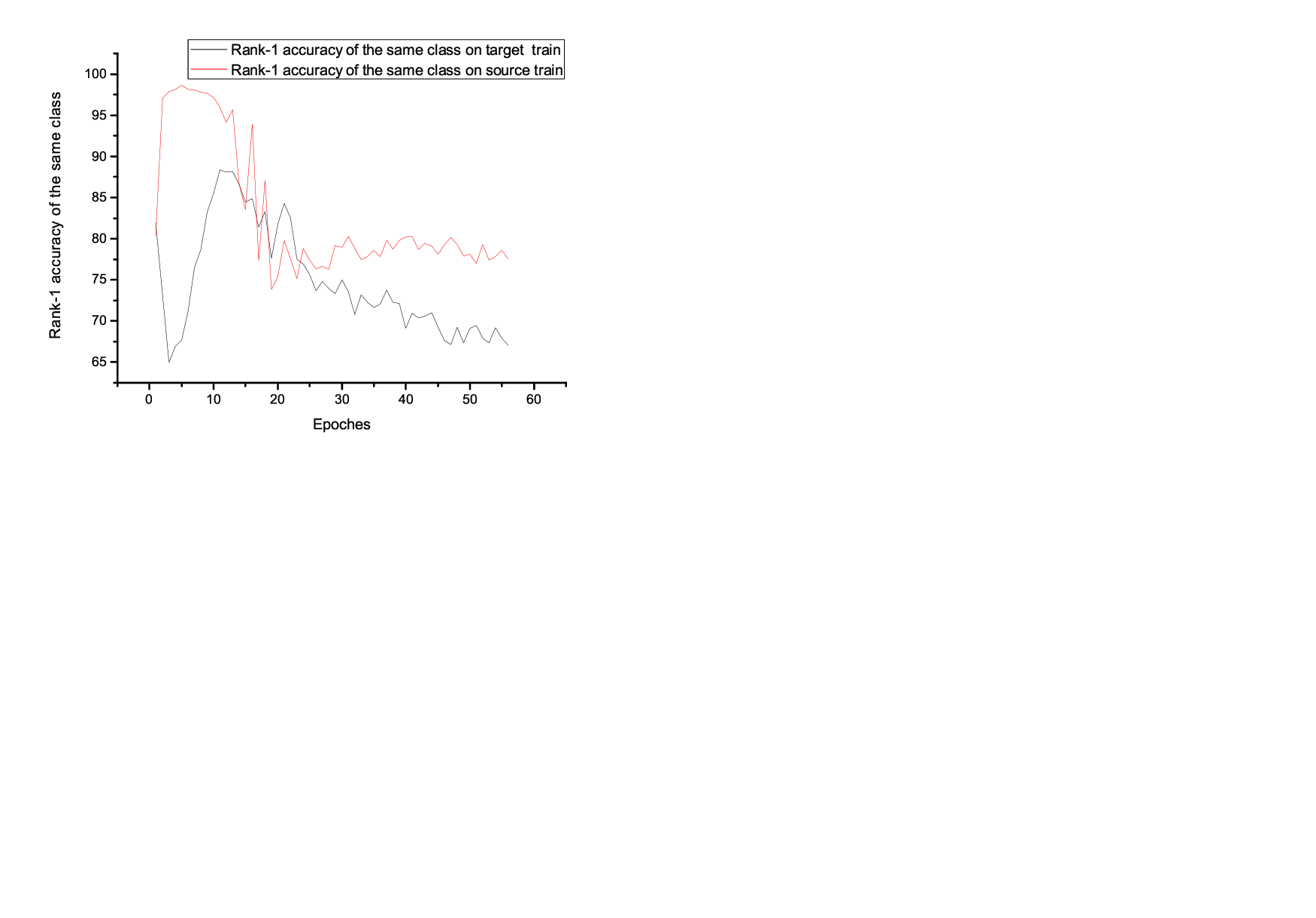}
     \vspace{-4cm}
    \caption{Source train and target train of the same class}
    \label{fig:d}
    \end{center} 
  \end{subfigure}
  \hfill
  \begin{subfigure}[b]{0.23\linewidth}
     \begin{center} 
     \includegraphics[scale=0.32]{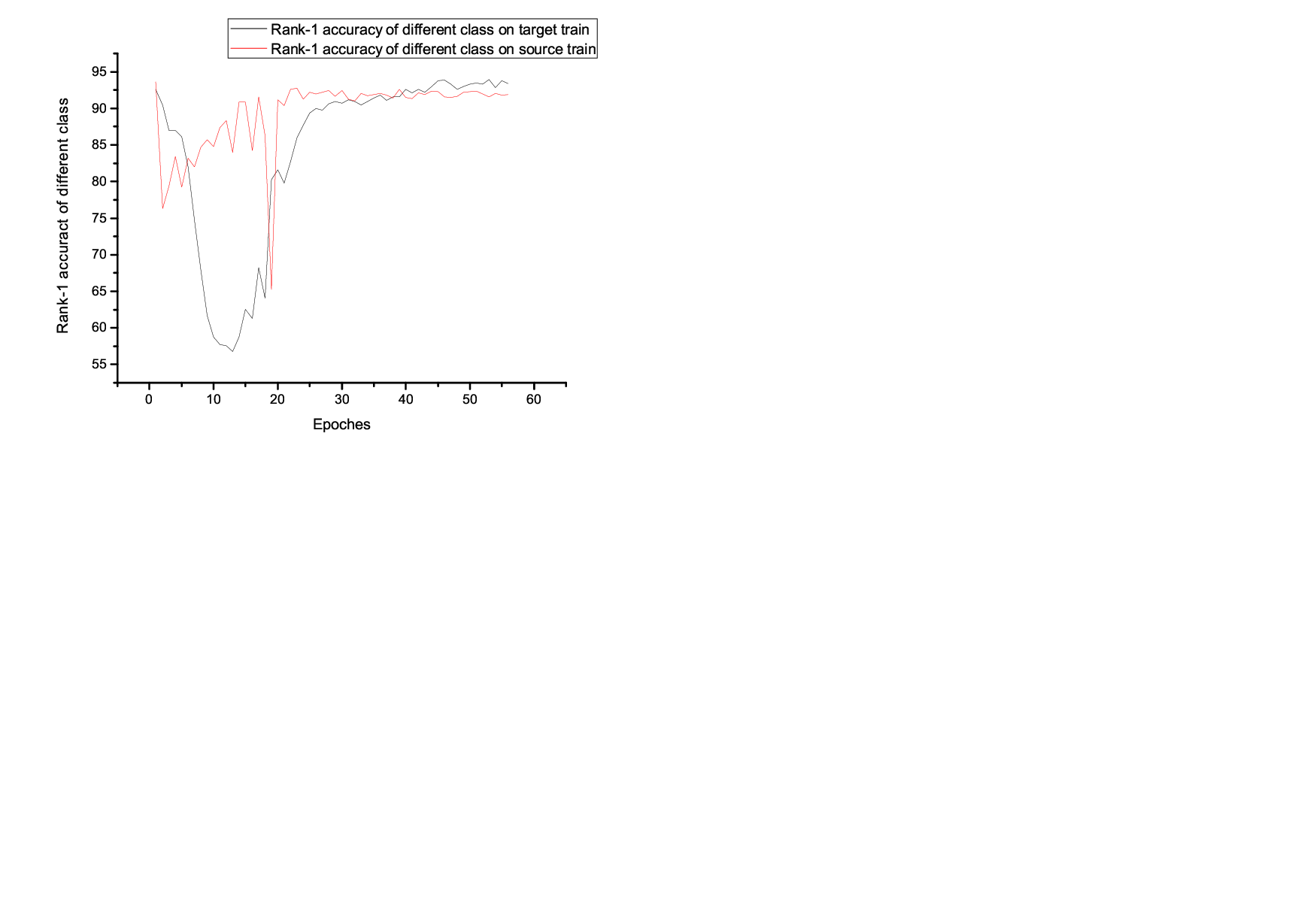}
     \vspace{-4cm}
    \caption{Source train and target train of the different class}
    \label{fig:e}
    \end{center} 
  \end{subfigure}
    \hfill
  \begin{subfigure}[b]{0.23\linewidth}
     \begin{center} 
     \includegraphics[scale=0.32]{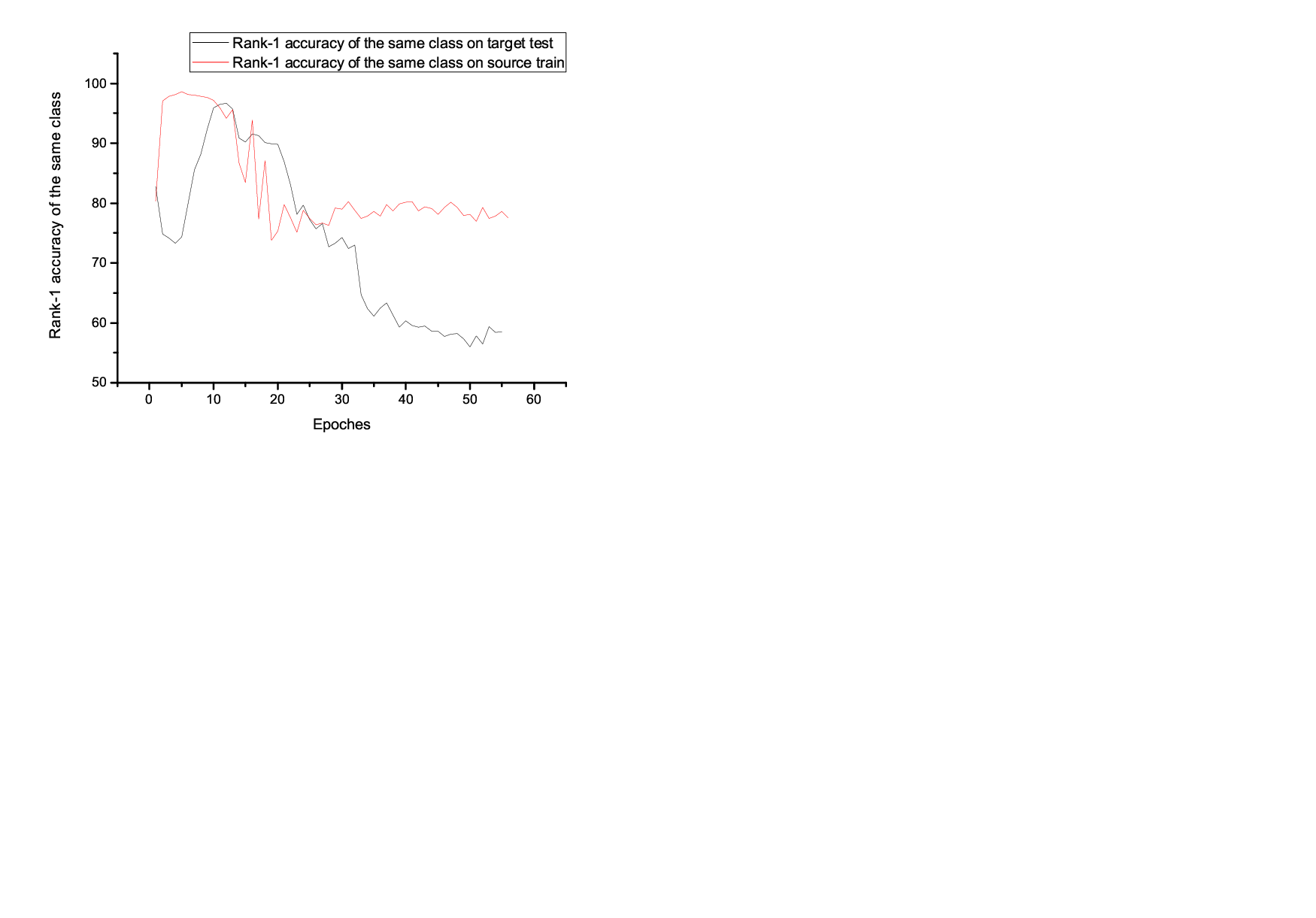}
     \vspace{-4cm}
    \caption{Source train and target test of the same class}
    \label{fig:f}
    \end{center} 
  \end{subfigure}
      \hfill
  \begin{subfigure}[b]{0.23\linewidth}
     \begin{center} 
     \includegraphics[scale=0.32]{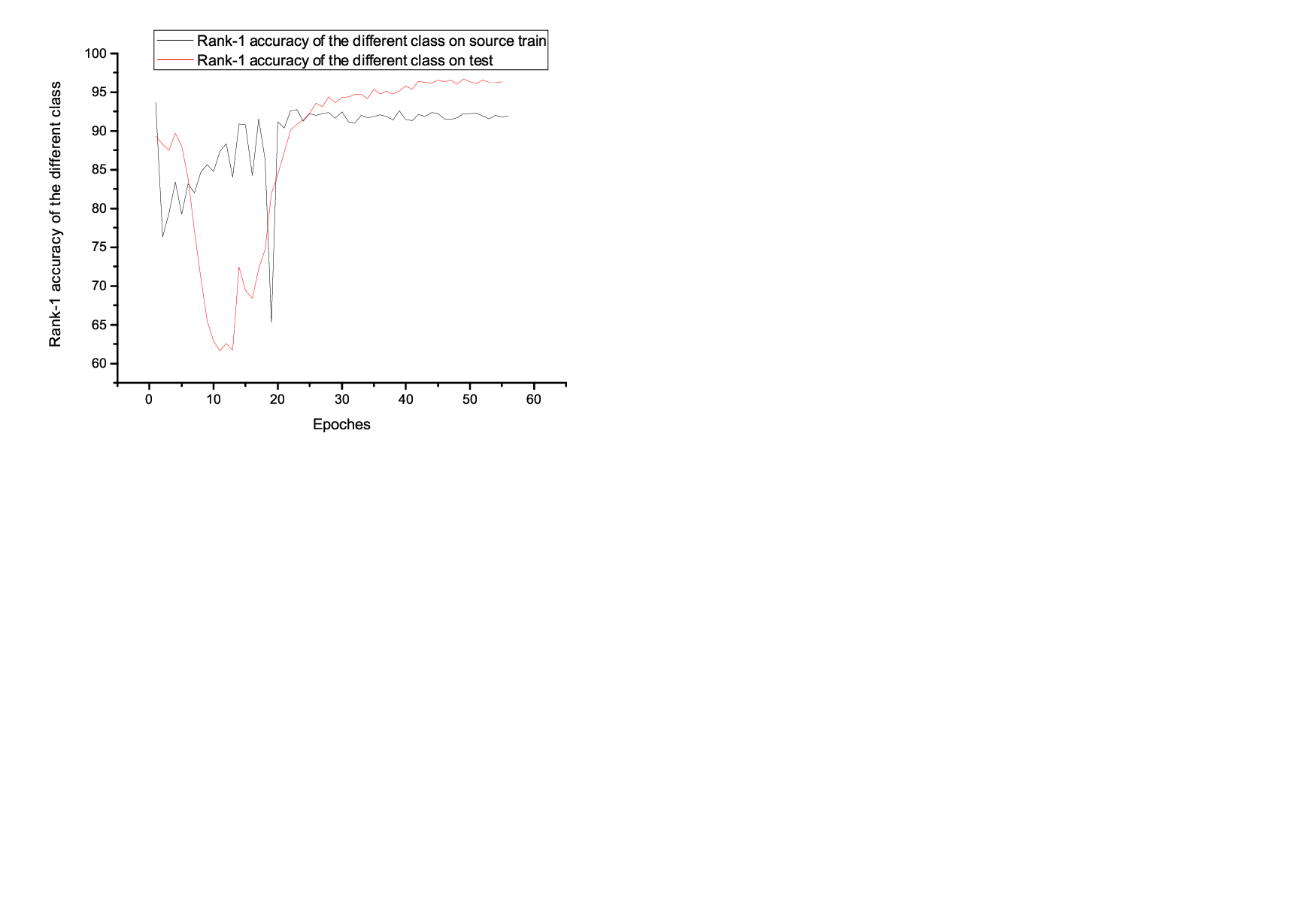}
     \vspace{-4cm}
    \caption{Source train and target test of the different class}
    \label{fig:g}
    \end{center} 
  \end{subfigure}
      \hfill
  \begin{subfigure}[b]{0.23\linewidth}
     \begin{center} 
     \includegraphics[scale=0.32]{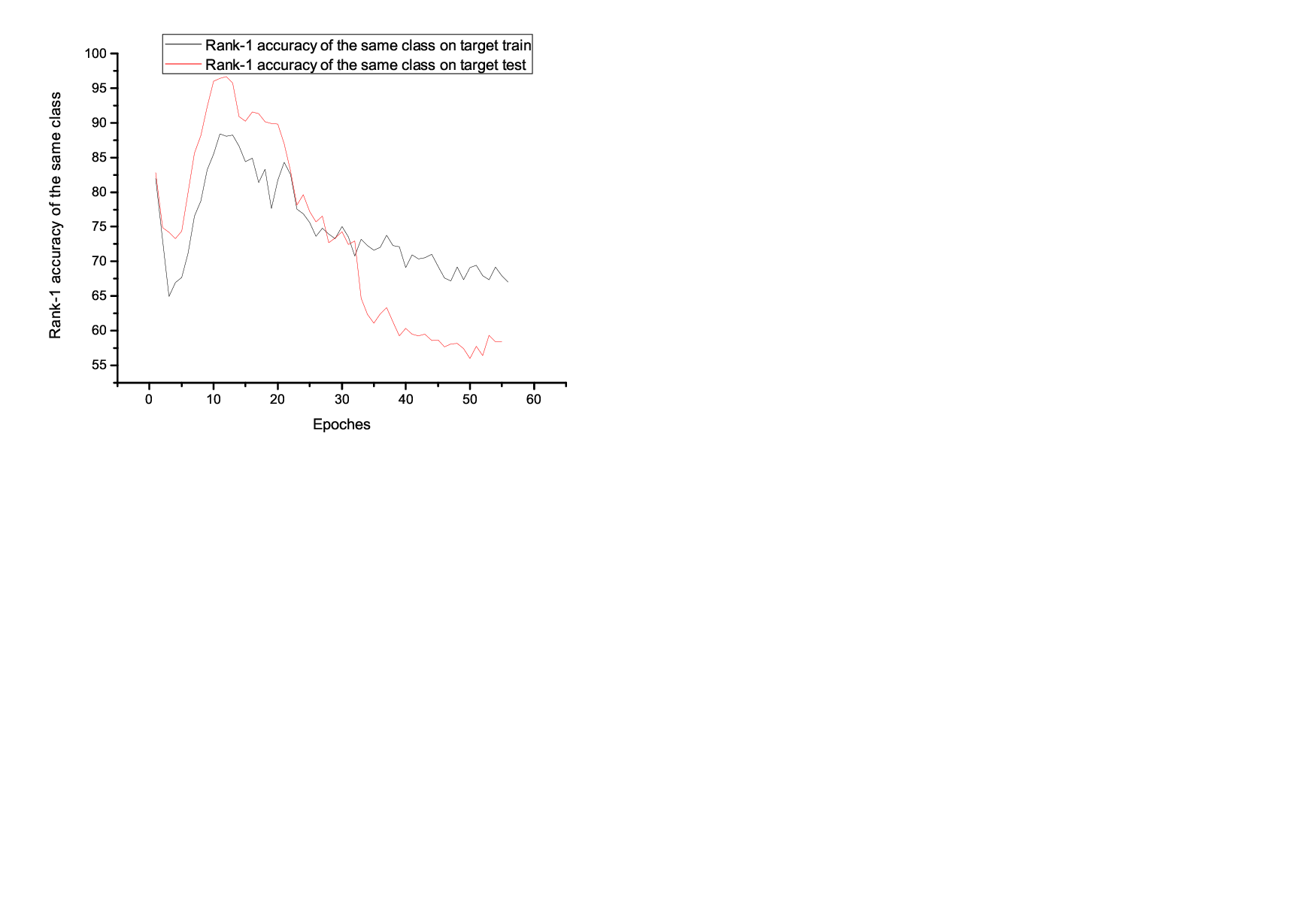}
     \vspace{-4cm}
    \caption{Target train and test of the same class}
    \label{fig:h}
    \end{center} 
  \end{subfigure}
      \hfill
  \begin{subfigure}[b]{0.23\linewidth}
     \begin{center} 
     \includegraphics[scale=0.32]{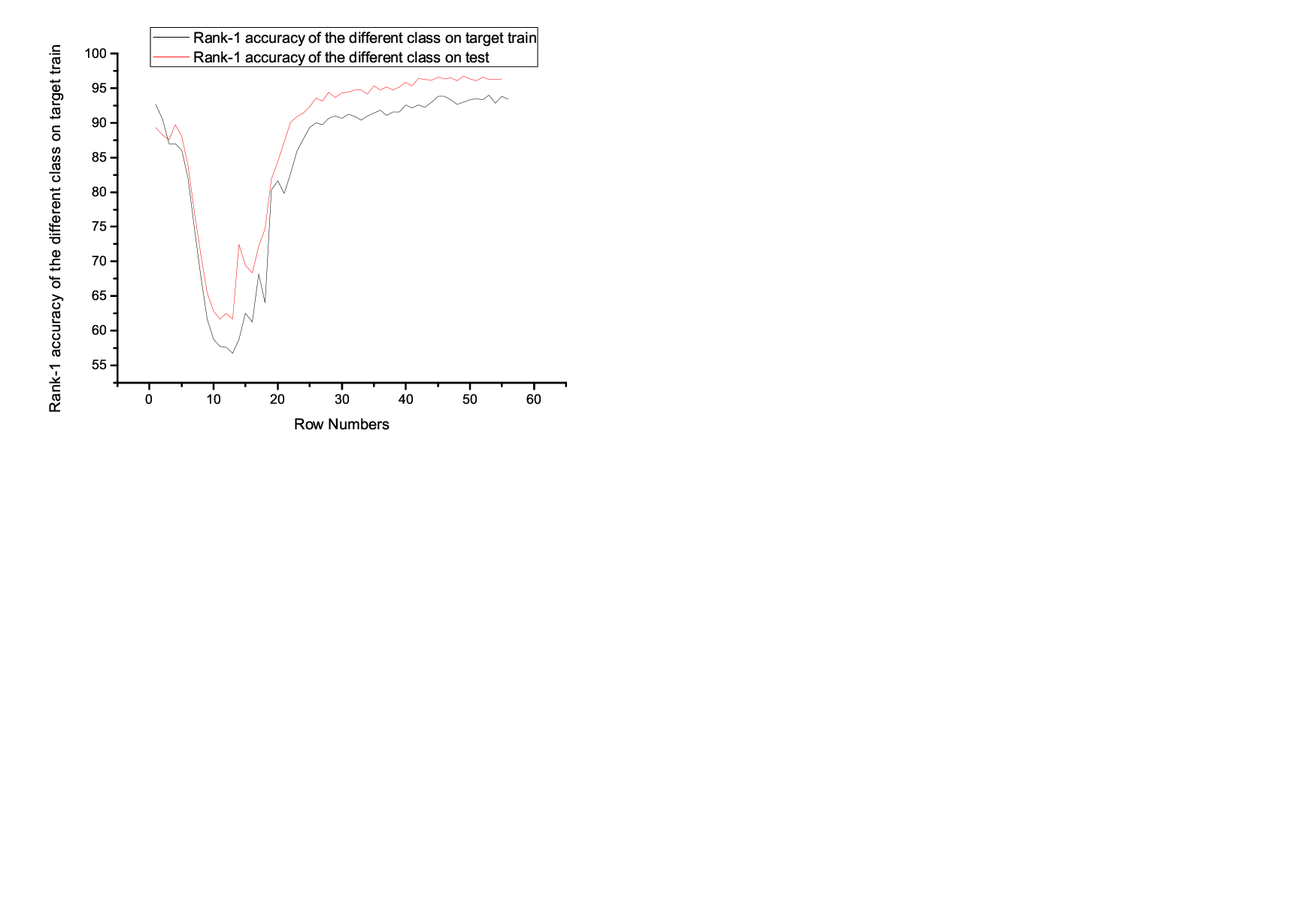}
     \vspace{-4cm}
    \caption{Target train and test of the different class}
    \label{fig:i}
    \end{center} 
  \end{subfigure}
        \hfill
    \begin{subfigure}[b]{0.23\linewidth}
     \begin{center} 
     \includegraphics[scale=0.32]{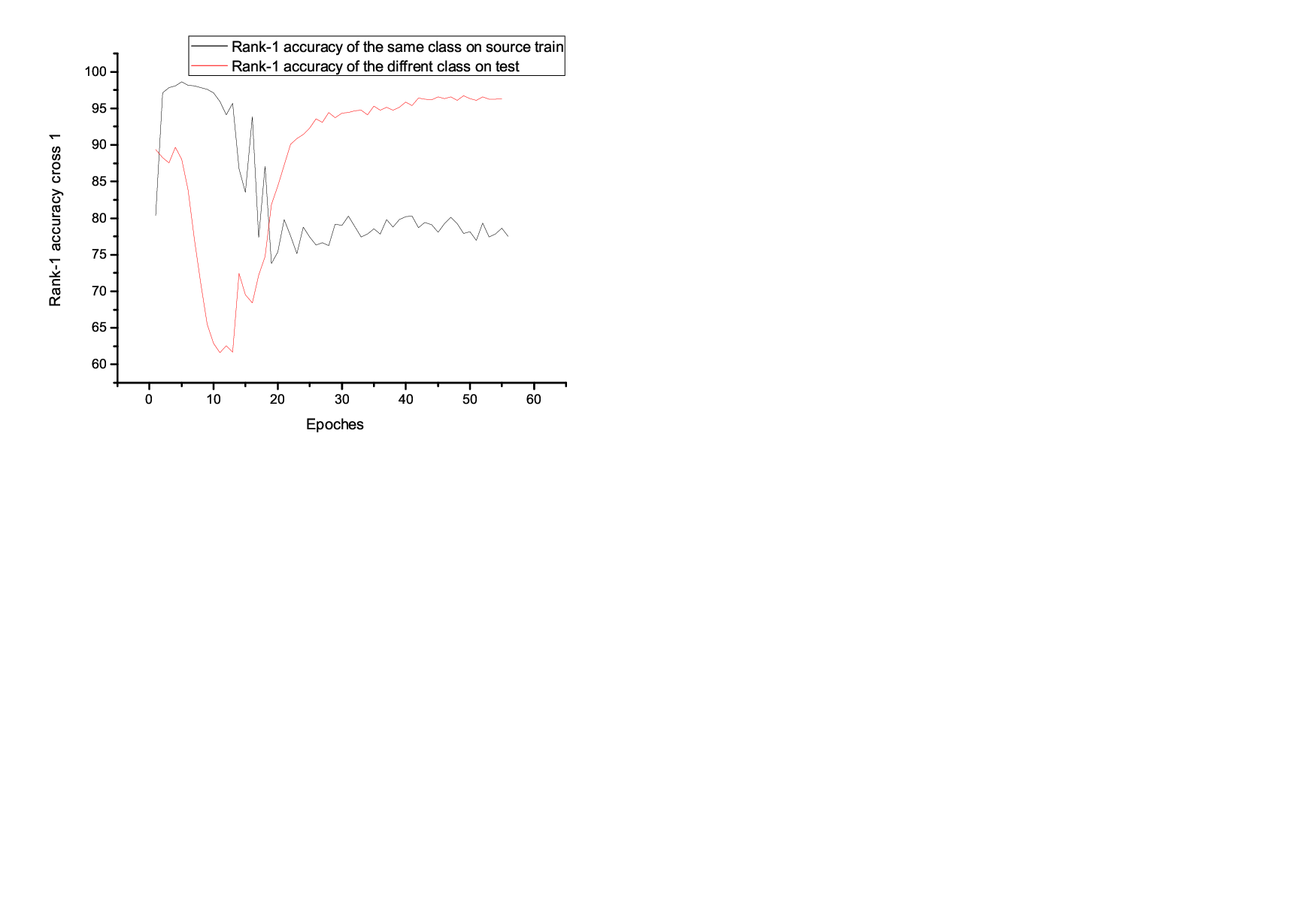}
     \vspace{-4cm}
    \caption{Source train and test of the corss type1}
    \label{fig:j}
    \end{center} 
  \end{subfigure}
          \hfill
    \begin{subfigure}[b]{0.23\linewidth}
     \begin{center} 
     \includegraphics[scale=0.32]{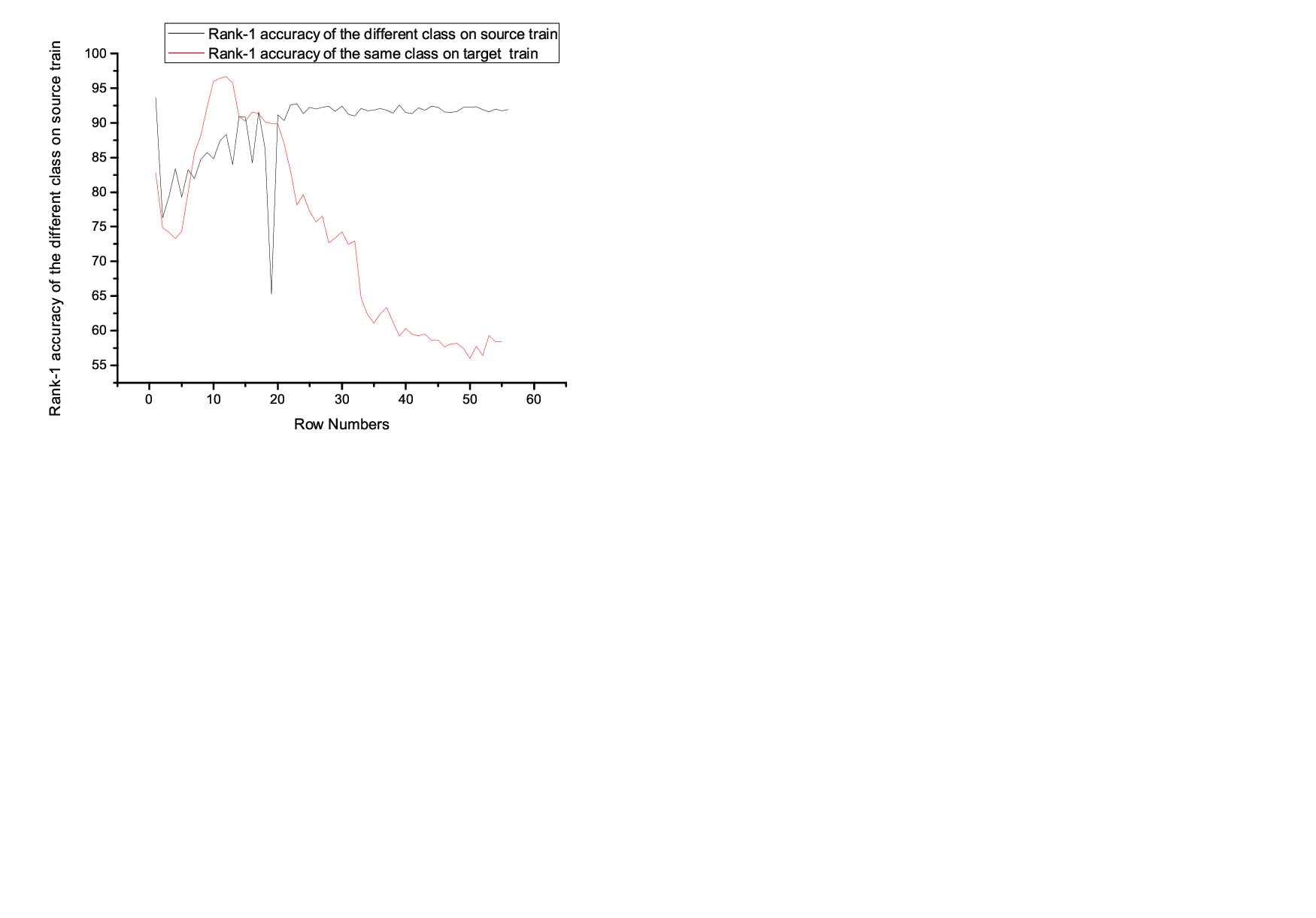}
     \vspace{-4cm}
    \caption{Source train and test of the corss type2}
    \label{fig:k}
    \end{center} 
  \end{subfigure}
          \hfill
    \begin{subfigure}[b]{0.23\linewidth}
     \begin{center} 
     \includegraphics[scale=0.32]{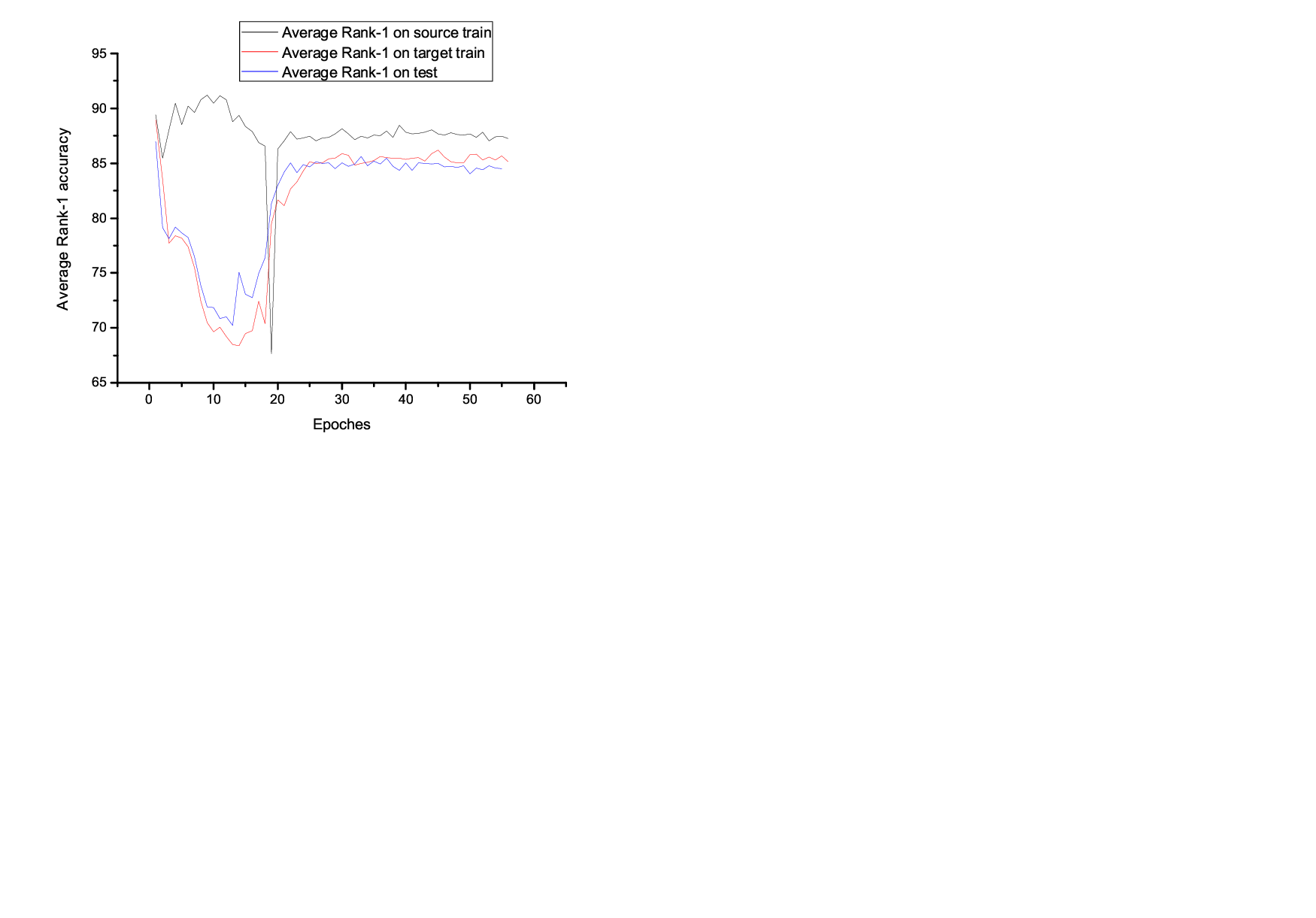}
     \vspace{-4cm}
    \caption{Average accuracy}
    \label{fig:l},
    \end{center} 
  \end{subfigure}
  \hfill
  \caption{Experiment evaluations}
  \label{fig06}
\end{figure*}
Figure \ref{fig06} is not a convergence curve but a wave function figure.
In this analogy, we can consider each epoch's model state as a state of the wave function, and calculate the probability of each state based on the model's accuracy on the validation set. 
States with higher amplitudes can be interpreted as having a higher probability. 
By analogy, we can consider that higher amplitude states correspond to more likely accuracies. 
Therefore, we conclude that the recognition rates of old knowledge ($97\%$) and new knowledge ($90\%$, Potential for higher performance observed) in the new world have the highest probabilities.
When the state and velocity of meta-learning quantum are influenced by learning rate updates (because Decays the learning rate of each parameter group by gamma every 20 epoches), the meta learning quantum are "observed," and one of the possible states of the particle becomes the actual state, with the probability of all other possible states abruptly transitioning to 0. 
The sudden and discontinuous change in the wave function observed at 20 epochs is referred to as "wave function collapse." 
Figure\ref{fig06} also demonstrates that the wave function of meta-learning quantum can be measured before collapse occurs.

From the comparison of the nine quantum systems in Figure \ref{fig06}, it can be seen that the quantum systems proposed in this paper have reached a state of quantum entanglement. 
The dependence between them is evident, and they cannot be independently described or decomposed into their individual states. 
Even if the quantum systems are physically or morphologically far apart in the old and new worlds, when two or more quantum systems are in an entangled state, their states are closely correlated.
The measurement results of one system immediately affect the states of other systems.
Furthermore, the presence of entangled states exhibits some non-classical characteristics in our quantum systems, such as parallel computing (as shown in Figure 6(d)(h)(i)), quantum teleportation (as shown in Figure 6(a)(b)(j)), and quantum error correction through entanglement (as shown in Figure 6(e)(f)(g)(l)).
Quantum channels are composed of model meta-features and parameters, and they are reconstructed by transmitting information through operations and measurement results of entangled pairs, providing security.

We also observed two interesting phenomena: 1.The learning ability of the new world has a traction effect on the learning ability of the old world, and the new world quantum systems can perform entanglement-based error correction on the old world quantum systems. 
Therefore, we are more certain that it is the entangled state between quantum systems that is playing a role.
2.Regardless of the new or old world, the ability to recognize new things and the ability to recognize old things always exhibit a trade-off. 
The reason may be that this study starts from essential cognition, similar to human cognition, where flexibility is needed and time is required to adjust when encountering new things, or old things are prone to be forgotten.

There is currently no previous successful work on quantum learning models.
According to reference \cite{f45}, currently, open-world models achieve less than $10\%$ accuracy on the open-set.
It is expected that the recognition accuracy will also be ineffective on the open-world test set.
Due to the presence of quantum superposition and quantum entanglement, similar to the glacier effect, in the first epoch, the cognitive level of the new world is already comparable to that of the old world and eventually reaching $97\%$. 
The cognitive level of the old world also quickly reaches $90\%$ from essential cognition and eventually reaching $97\%$.
This demonstrates that our quantum system exhibits excellent performance in recognizing the new world, and it is particularly valuable that it possesses the error correction capability of the old world's experience.

\section{Conclusion}\
This paper focuses on addressing the biggest challenge in the process of AI exploring new knowledge: the inconsistency in measuring and units of features between the new and old worlds. 
Within the framework of semi-supervised learning in the new and old worlds, the evolutionary direction of feature learning in the new world is guided by the meta-characteristic of both worlds. 
By learning the meta-characteristic of the new and old worlds, the knowledge discriminator of the new and old worlds is guided by the evolution of elemental features, forming the learning capability baseline for the new and old worlds.
As elemental features are adept at capturing the feature symmetry between the new and old worlds, the proposed model system enhances the experiential transformation of existing feature knowledge. 
Through testing on pedestrian re-identification datasets, we have already observed a recognition accuracy of $97\%$ in the unknown world. 
We may need to understand whether AI will surpass humans in discovering new knowledge.
It is worth emphasizing that the system provided in this paper is a metaphorical extension of quantum concepts to learning systems and should not be regarded as a literal description of a clearly defined field. 
The application of quantum principles to learning systems is an area of ongoing research and exploration. The actual impact and methodologies are still evolving and under investigation.
In the future, we will introduce real quantum computation, the amplitudes and probability distributions of a wave function are determined through quantum superposition and interference, rather than through classical statistical methods.


%

\appendices
%
%


\ifCLASSOPTIONcaptionsoff
\newpage
\fi



%

%


\bibliographystyle{IEEEtran}
\bibliography{RefJWang}

\begin{thebibliography}{10}
\providecommand{\url}[1]{#1}
\csname url@samestyle\endcsname
\providecommand{\newblock}{\relax}
\providecommand{\bibinfo}[2]{#2}
\providecommand{\BIBentrySTDinterwordspacing}{\spaceskip=0pt\relax}
\providecommand{\BIBentryALTinterwordstretchfactor}{4}
\providecommand{\BIBentryALTinterwordspacing}{\spaceskip=\fontdimen2\font plus
\BIBentryALTinterwordstretchfactor\fontdimen3\font minus
  \fontdimen4\font\relax}
\providecommand{\BIBforeignlanguage}[2]{{%
\expandafter\ifx\csname l@#1\endcsname\relax
\typeout{** WARNING: IEEEtran.bst: No hyphenation pattern has been}%
\typeout{** loaded for the language `#1'. Using the pattern for}%
\typeout{** the default language instead.}%
\else
\language=\csname l@#1\endcsname
\fi
#2}}
\providecommand{\BIBdecl}{\relax}
\BIBdecl

\bibitem{f44}
W.~J. Scheirer, A.~de~Rezende~Rocha, A.~Sapkota, and T.~E. Boult, ``Toward open
  set recognition,'' \emph{IEEE transactions on pattern analysis and machine
  intelligence}, pp. 1757--1772, 2012.

\bibitem{f45}
A.~Bendale and T.~Boult, ``Towards open world recognition,'' pp. 1893--1902,
  2015.

\bibitem{f12}
F.~Chen, N.~Wang, J.~Tang, P.~Yan, and J.~Yu, ``Unsupervised person
  re-identification via multi-domain joint learning,'' \emph{Pattern
  Recognition}, p. 109369, 2023.

\bibitem{f4}
J.~Sun, Y.~Li, H.~Chen, Y.~Peng, and J.~Zhu, ``Unsupervised cross domain person
  re-identification by multi-loss optimization learning,'' \emph{IEEE
  Transactions on Image Processing}, pp. 2935--2946, 2021.

\bibitem{f14}
H.-X. Yu, W.-S. Zheng, A.~Wu, X.~Guo, S.~Gong, and J.-H. Lai, ``Unsupervised
  person re-identification by soft multilabel learning,'' \emph{Proceedings of
  the IEEE/CVF conference on computer vision and pattern recognition}, pp.
  2148--2157, 2019.

\bibitem{f30}
``Controllable person image synthesis with attribute-decomposed gan,''
  \emph{Men, Yifang and Mao, Yiming and Jiang, Yuning and Ma, Wei-Ying and
  Lian, Zhouhui}, pp. 5084--5093, 2020.

\bibitem{f7}
P.~Peng, T.~Xiang, Y.~Wang, M.~Pontil, S.~Gong, T.~Huang, and Y.~Tian,
  ``Unsupervised cross-dataset transfer learning for person
  re-identification,'' pp. 1306--1315, 2016.

\bibitem{f11}
M.~Ghifary, W.~B. Kleijn, M.~Zhang, D.~Balduzzi, and W.~Li, ``Deep
  reconstruction-classification networks for unsupervised domain adaptation,''
  pp. 597--613, 2016.

\bibitem{f48}
W.~Liu, Y.~Wen, Z.~Yu, and M.~Yang, ``Large-margin softmax loss for
  convolutional neural networks,'' \emph{arXiv preprint arXiv:1612.02295},
  2016.

\bibitem{f49}
Y.~Tang, Z.~Zhao, C.~Li, and X.~Ye, ``Open set recognition algorithm based on
  conditional gaussian encoder,'' \emph{Mathematical Biosciences and
  Engineering}, pp. 6620--6637, 2021.

\bibitem{f46}
A.~Bendale and T.~E. Boult, ``Towards open set deep networks,''
  \emph{Proceedings of the IEEE conference on computer vision and pattern
  recognition}, pp. 1563--1572, 2016.

\bibitem{f10}
H.~Chen, Y.~Wang, Y.~Shi, K.~Yan, M.~Geng, Y.~Tian, and T.~Xiang, ``Deep
  transfer learning for person re-identification,'' \emph{2018 IEEE Fourth
  International Conference on Multimedia Big Data (BigMM)}, pp. 1--5, 2018.

\bibitem{f23}
S.~J. Pan and Q.~Yang, ``A survey on transfer learning,'' \emph{IEEE
  Transactions on knowledge and data engineering}, pp. 1345--1359, 2009.

\bibitem{f5}
J.~Lv, W.~Chen, Q.~Li, and C.~Yang, ``Unsupervised cross-dataset person
  re-identification by transfer learning of spatial-temporal patterns,'' pp.
  7948--7956, 2018.

\bibitem{f24}
A.~Farahani, S.~Voghoei, K.~Rasheed, and H.~R. Arabnia, ``A brief review of
  domain adaptation,'' \emph{Advances in data science and information
  engineering: proceedings from ICDATA 2020 and IKE 2020}, pp. 877--894, 2021.

\bibitem{f8}
A.~Hermans, L.~Beyer, and B.~Leibe, ``In defense of the triplet loss for person
  re-identification,'' \emph{arXiv preprint arXiv:1703.07737}, 2017.

\bibitem{f31}
J.~Wang, Y.~Song, T.~Leung, C.~Rosenberg, J.~Wang, J.~Philbin, B.~Chen, and
  Y.~Wu, ``Learning fine-grained image similarity with deep ranking,''
  \emph{Proceedings of the IEEE conference on computer vision and pattern
  recognition}, pp. 1386--1393, 2014.

\bibitem{f32}
A.~Gretton, K.~M. Borgwardt, M.~J. Rasch, B.~Sch{\"o}lkopf, and A.~Smola, ``A
  kernel two-sample test,'' \emph{The Journal of Machine Learning Research},
  pp. 723--773, 2012.

\bibitem{f9}
F.~Schroff, D.~Kalenichenko, and J.~Philbin, ``Facenet: A unified embedding for
  face recognition and clustering,'' \emph{Proceedings of the IEEE conference
  on computer vision and pattern recognition}, pp. 815--823, 2015.

\bibitem{f50}
R.~Chrisley, ``Quantum learning,'' \emph{New directions in cognitive science:
  Proceedings of the international symposium, Saariselka}, vol.~4, 1995.

\end{thebibliography}
\end{document}